\documentclass{article}
\usepackage{spconf,amsmath,epsfig}

\usepackage{url}
\usepackage{graphicx}
\usepackage{epstopdf}
\usepackage{caption}
\usepackage{subcaption}
\usepackage{booktabs}
\usepackage{color}
\usepackage{multirow}
\usepackage{makecell}
\usepackage{bm}
\usepackage{hyperref}
\usepackage{amssymb}
\usepackage{adjustbox}
\usepackage{enumitem}
\usepackage[compress]{cite}
\newcommand{\etal}{\textit{et al.}}

\let\OLDthebibliography\thebibliography
\renewcommand\thebibliography[1]{
  \OLDthebibliography{#1}
  \setlength{\parskip}{0pt}
  \setlength{\itemsep}{0pt plus 0.3ex}
}

\begin{document}\sloppy

\def\x{{\mathbf x}}
\def\L{{\cal L}}

\title{Point Cloud Compression via Constrained Optimal Transport}
%

\name{Zezeng Li$^{1}$ \qquad Weimin Wang$^{1}$ \qquad Ziliang Wang \qquad Na Lei$^{1 \star}$}

\address{$^{1}$ School of Software, Dalian University of Technology, Dalian, 116024, People's Republic of China}

\maketitle

\begin{abstract}
This paper presents a novel point cloud compression method {COT-PCC} by formulating the task as a constrained optimal transport~(COT) problem. COT-PCC takes the bitrate of compressed features as an extra constraint of optimal transport~(OT) which learns the distribution transformation between original and reconstructed points. 
Specifically, the formulated COT is implemented with a generative adversarial network~(GAN) and a bitrate loss for training. The discriminator measures the Wasserstein distance between input and reconstructed points, and a generator calculates the optimal mapping between distributions of input and reconstructed point cloud. Moreover, we introduce a learnable sampling module for downsampling in the compression procedure. Extensive results on both sparse and dense point cloud datasets demonstrate that COT-PCC outperforms state-of-the-art methods in terms of both CD and PSNR metrics. Source codes are available at \url{https://github.com/cognaclee/PCC-COT}.
\end{abstract}

\section{Introduction}
\label{sec:intro}
With the wide deployment of 3D sensors in autonomous vehicles and robot systems, millions of points are generated within mere seconds. While it is worth celebrating that complex geometry can now be captured with fine details, significant challenges in data transmission and storage are also introduced. 
As a consequence, point cloud compression (PCC) becomes imperative for the broader applicability. Traditional PCC algorithms usually apply image/video compression technology to range images derived by 3D data~(e.g., V-PCC~\cite{graziosi2020overview,zhang2022novel} or utilize more efficient data structures (e.g., G-PCC or octree-based~\cite{huang2020octsqueeze}). 

Recently, inspired by the success of deep learning in image compression and point cloud analysis, researchers also pay attention to learning-based PCC approaches. Depending on the representation of point cloud data, existing learning-based methods can be categorized into voxel-based~\cite{quach2019learning,wang2021lossy,quach2020improved,wang2022sparse,song2023efficient,liu2023pchm} and point-based  ones\cite{huang20193d,yan2019deep,gu20203d,wiesmann2021deep, wang2021multiscale,He_2022_CVPR,gao2023point}.
\textbf{Voxel-based} methods divide point cloud into organized grid structures which often leads to excessive complexity increment in time and memory as the resolution increases. Considering this, Que \etal~\cite{que2021voxelcontext}, Wang \etal~\cite{wang2022sparse} and Liu \etal~\cite{liu2023pchm} propose to improve efficiency with sparse tensors and octrees. In contrast, \textbf{point-based} methods directly consume 3D discrete points, which enables the preservation of fine local geometric details. 
Among them, 
a point-based auto-encoder is utilized to achieve compression and decompression~\cite{huang20193d,yan2019deep}.
Subsequently, multiscale schemes and downsample/upsample strategies are also proposed~\cite{wiesmann2021deep,wang2021multiscale}.
While these methods achieve remarkable performance in compressing uniform data, they may fail to preserve various densities for practically acquired non-uniform point clouds. 
Noticing this problem, He \etal~\cite{He_2022_CVPR} propose to encode local geometry with density-aware hierarchical embedding by introducing density and cardinality losses. However, excessive focus on local density may lead to suboptimal performance due to the underestimation of global distribution. On the other hand, the Rate-Distortion-Perception~(RDP) model~\cite{2019Rethinking,yan2021perceptual,zhang2021universal} is proposed for image compression with high perceptual quality by incorporating global distribution constraints.
Nonetheless, the non-Euclidean structure of point clouds and the change of the metric from Euclidean distance to geodesic distance prevent these models from being directly applied to PCC.

To preserve both local density and global distribution, we propose \textbf{COT-PCC} by formulating PCC as a constrained optimal transport~(COT) problem. 
The key idea is to learn an optimal representation for distribution mapping while constraining the bitrate. We first design a network suitable for non-Euclidean data to parameterize the mapping, and then introduce a quadratic Wasserstein distance to accurately measure the distance between two point clouds to guide model learning.
Moreover, a learnable sampler is introduced to better fit COT in point selection for compression, which is different from the existing Farthest Point Sampling~(FPS).
Contributions of this paper are three-fold: 

\begin{itemize}[noitemsep]
\vskip -0.3in
\item 
We innovatively formulate PCC tasks as the COT problem. Consequently, better compression performance can be achieved by solving the COT problem. 
\item 
We introduce a learnable sampler to facilitate the downsampling stage of the compression process
by learning to select
points that are beneficial to the compression.
\item 
We implement COT-PCC with GANs and perform extensive experiments on both sparse and dense point clouds to validate the advantages of COT-PCC. 
\end{itemize}

\vskip -0.1in
\begin{figure*}[!t]
    \begin{center}
    \includegraphics[width=0.95\linewidth]{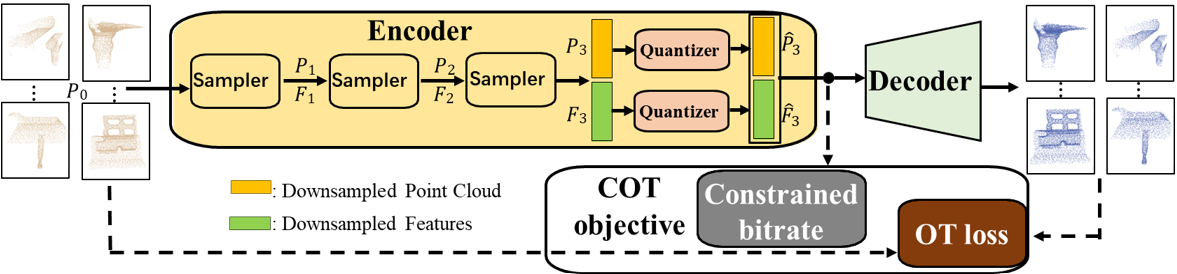}
    \caption{The framework of the proposed \textbf{COT-PCC}. The Encoder consists of three stages of Sampler, which is a learnable sampling module. As the output of the last stage,
    coordinates $P_3$ and features $F_3$ are followed by Quantizer for bit compression. Decoder reconstructs the data from compressed $\hat{P}_3$ and $\hat{F}_3$. The COT objective quantifies the reconstruction performance via OT loss within the bitrate constraint.
    Dash lines indicate the process only in the training phase.
    }
    \label{Framework}
    \end{center}
    \vskip -0.3in
\end{figure*}
\section{Methodology}
In PCC tasks, low bitrate and loss of distribution alignment are two main objectives. PCC can be modeled as rate-distortion~(RD) models~\cite{de2016compression,quach2020improved, biswas2020muscle, quach2019learning,wang2021lossy, que2021voxelcontext,wang2021multiscale,He_2022_CVPR,gao2023point,guarda2020adaptive,2019Rethinking,yan2021perceptual,zhang2021universal}, which achieve the tradeoff between bitrate and distortion. 
To also align the global distribution with reconstructed data under restrained bitrate, we propose to combine RD and OT
,
thus converting the PCC into a COT problem. 
We first briefly introduce the preliminaries of RD and OT, whose combination is utilized as local and global constraints for point cloud compression.
\subsection{Preliminaries}
\noindent\textbf{Rate-Distortion Model.}
Rate–distortion model (RD)~\cite{thomas2006elements} characterizes the fundamental tradeoff between the bitrate and the distortion. The Lagrangian formulation of the RD model is given by $R(D) = D + \lambda R$, where $D$ penalizes distortion and $R$ penalizes bitrate, $\lambda$ is used to balance bitrate and distortion. The RD model is a convex and non-increasing function of $D$, demonstrating the rate-distortion tradeoff. Under different information metrics, different RD models can be obtained. Numerous studies~\cite{shannon1959coding,balle2016end,yan2021perceptual,liu2021lossy,quach2019learning,wen2020lossy} employ Shannon entropy 
$H(\cdot)$ as the measure of information for compressed output $Z$. 
Then RD model becomes:
\begin{equation}\label{eq:RDH}
	R^{(H)}(D) = D + \lambda H(Z).
\end{equation}

 \noindent\textbf{Optimal Transport.}
The OT problem seeks to identify the most efficient mapping between two distributions while minimizing the transportation cost, which has been widely applied in machine vision and machine learning~\cite{peyre2019computational,li2022weakly}. 
OT problem was initially introduced in the context of Monge's problem~\cite{monge1781memoire} described as follows.

\noindent\textbf{Monge's problem}: Suppose $\mu  \sim {\cal P}(X)$and $\nu  \sim {\cal P}(Y)$ be two sets of  probability measures defined on $X$ and $Y$, respectively, and cost function $c(x,y):X \times Y \to [0, + \infty ]$ measure the cost of transporting $x \in X$ to $y \in Y$. 
To find a mapping $T:X \to Y$ to turn the mass of $\mu$ into $\nu$ by
\begin{equation}
    \mathop {\inf }\limits_T \int_X {c\left( {x,T(x)} \right)} d\mu (x)\qquad  {\rm{subject \ to\ }}~~\nu  = {T_\# }\mu ,
\label{eq:Monge}
\end{equation}
where ${T_\# }\mu$ is the push-forward measure induced by $T$. 
Correspondingly, the minimal distance between the $X$ and $Y$ is the Wasserstein distance.

\subsection{Formulation of PCC with COT}
In the process  $X \xrightarrow{f_E} Z \xrightarrow{f_D} \hat X$, we denote 
compressed output as $Z$ by encoder $f_E$ and $\hat X$ as reconstructed data by the decoder $f_D$. Then, the discrete COT problem is defined as:\\
\textbf{Definition 1. Constrained optimal transport problem}:
\begin{equation}
	\begin{gathered}
		\mathop {\min }\limits_{T} {\mathbb{E}_{X \sim {p_X}}}\left[ c(X,T(X))\right]  \\ 
		{\text{subject \ to \ }}~~\ {p_{\hat X}} = {p_X},H(Z)\leq R,
	\end{gathered}
	\label{eq:CMonge}
\end{equation}
where $c(X,T(X))$ represents $D$ in Eq.\eqref{eq:RDH}, $T(X)=f_D\circ f_E(X)$ denotes the composition of $f_D$ and $f_E$. Eq.\eqref{eq:CMonge} is a constrained Monge's OT problem with the constraints of bitrate $H(Z)\leq R$ of Eq.\eqref{eq:RDH}. For ease of implementation, 
we use the Lagrange multiplier method to relax Eq.\eqref{eq:CMonge} into an unconstrained one as
\begin{equation}
	\mathop {\min }\limits_{T} {\mathbb{E}_{X \sim {p_X}}}[ c(X,T(X)) + \beta d({p_X},{p_{\hat X}}) + \lambda H(Z)],
	\label{eq:COT}
\end{equation}
where $d({p_X},{p_{\hat X}})$ measures the deviation between distributions ${p_X}$ and the ${p_{\hat X}}$, and $\beta > 0$ is a balance parameter. Under mild conditions, Eq.\eqref{eq:COT} with a penalty parameter $\beta $ is equivalent to the formulation with a constraint $d({p_X},{p_{\hat X}}) \le {\mu _\beta}$ for some ${\mu _\beta} > 0$. As $\beta $ increases, the value of ${\mu _\beta }$ decreases. As $\beta  \to \infty $, the solution of Eq.\eqref{eq:COT} satisfies that $d({p_X},{p_{\hat X}}) \to 0$, i.e. ${p_{\hat X}} \to {p_X}$. More generally, Eq.\eqref{eq:Monge} serves as a lower bound for Eq.\eqref{eq:COT} for any $R > 0$. 

As the combination of OT and RD, Eq.\eqref{eq:COT} simultaneously achieves three objectives: local distortion including density constrained by $c(X,T(X)$, global distribution constrained by $d({p_X},{p_{\hat X}})$, and the bitrate of $Z$ constrained by $R$.
\subsection{Implementation of COT}
COT-PCC is designed to simultaneously focus on aligning with the global distribution and effectively preserving the local density of the point cloud. The preservation of global distribution is
primarily achieved through the design of the objective, while the preservation of local features is achieved through the design of the objective and networks. Before explaining the algorithm, we first summarize the above analysis and give the following remark.\\
\textbf{Remark 1.} \textit{Let $X$, $Z$, $\hat X$ denote the source, compressed, and reconstructed point cloud set. $X$ is the input of the encoder $f_E$ and satisfies $X\sim{p_X}$, $\hat X$ is the output of the decoder $f_D$ and satisfies $\hat X\sim{p_{\hat X}}$. The optimal encoder and decoder to Eq.\eqref{eq:COTPCC} is also a solution to COT problem Eq.\eqref{eq:COT}}.
\begin{equation}
	\begin{aligned}
		\mathop {\min }\limits_{f_D, f_E} {\mathbb{E}_{X \sim {p_X}}}[ c(X,\hat X) + 
		\beta d_{wass}({p_X},{p_{\hat X}}) + \lambda H(f_E(X))],
		\label{eq:COTPCC}
	\end{aligned}
\end{equation}
where $H(\cdot)$ denotes the Shannon entropy, $d_{wass}({p_X},{p_{\hat X}})$ is the Wasserstein distance between $X$ and $\hat X$.

\textbf{OT Loss.} The next question is, how do we calculate the Wasserstein distance $d_{wass}({p_X},{p_{\hat X}})$ and OT mapping $T(X)=f_D\circ f_E(X)$ in the PCC task? As mentioned in~\cite{lei2019geometric,li2022weakly}, GANs accomplish two major tasks: the generator computes the OT mapping, while the discriminator computes the Wasserstein distance. GAN-based methods utilize a discriminator to evaluate the point sets produced by the generator, which can help the generator to produce a rich variety of output patterns and regularize the reconstructed points from a global perspective~\cite{li2019pu,2018Unreasonable,yan2021perceptual}. Furthermore, Blau~\etal~\cite{2019Rethinking}stated that GANs can minimize the $L^1$ Wasserstein distance between the reconstructed distribution and the target distribution, thereby enhancing the perceptual quality of reconstructed images in image compression tasks. Here, considering the faster and more stable convergence efficiency of $L^2$ compared to $L^1$\cite{wganqc}, 
we adopt a GAN-based model for the PCC and train a discriminator $J$ to calculate quadratic Wasserstein $d_{wass}({p_X},{p_{\hat X}})$:
\begin{equation}	 
d_{wass}({p_X},{p_{\hat X}})=(J(X)-J(\hat X))^2.
\label{Wasserstein}
\end{equation}

Since infinite solutions exist for Eq. (\ref{eq:COTPCC}), we adopt an OT regularizer $L_{OTR}$ similar to ~\cite{wganqc,li2022weakly} to constrain the solution space of $J$ and stabilize GAN's training, defined as follows:
\begin{equation}	 
L_{OTR}={\mathbb{E}_{X \sim {p_X}}}[(||\nabla_xJ\left(X\right)|| - c(X,\hat X))^2]\ .
\label{OTR}
\end{equation}
Where $\nabla_xJ$ is the derivative of $J$ w.r.t $x$. For the distortion
measure $c(X,\hat X)$, we choose the $L^2$ Chamfer distance~\cite{huang2020octsqueeze}. 
After adding the regular term $L_{OTR}$, we optimize the encoder $f_E$, decoder $f_D$, and discriminator $J$ through adversarial training in Eq.\eqref{eq:COT-loss}, where $L_{OT}$ denotes the OT loss.
\begin{equation}
    L_{OT} = c(X,\hat X) + \beta d_{wass}({p_X},{p_{\hat X}})+\gamma L_{OTR}
    \label{eq:PCCGAN}
\end{equation}

\textbf{COT loss.}
Finally, the desired encoder-decoder pair $(f_E,f_D)$ is obtained by a training procedure with alternate optimization: \textit{i) The encoder and decoder $(f_E,f_D)$ are optimized by minimizing Eq.\eqref{eq:COT-loss}.} \textit{ii) The discriminator $J$ is optimized by maximizing Eq.\eqref{eq:COT-loss}}. Eq.\eqref{eq:COT-loss} guides the optimization of $f_E$ and $f_D$ in terms of global distribution, local density, and bitrate, respectively. This enables the proposed COT-PCC to simultaneously maintain good global distribution and local density under the given bit constraint.
\begin{equation}
    \mathop {\min }\limits_{f_E,f_D}\mathop{\max}\limits_{J} \ {\mathbb{E}_{X \sim {p_X}}}[ L_{OT} + \lambda H(f_E(X))]\\
    \label{eq:COT-loss}
\end{equation}

\begin{figure*}[htbp]
    \begin{center}	
    \centerline{\includegraphics[width=1.0\linewidth]{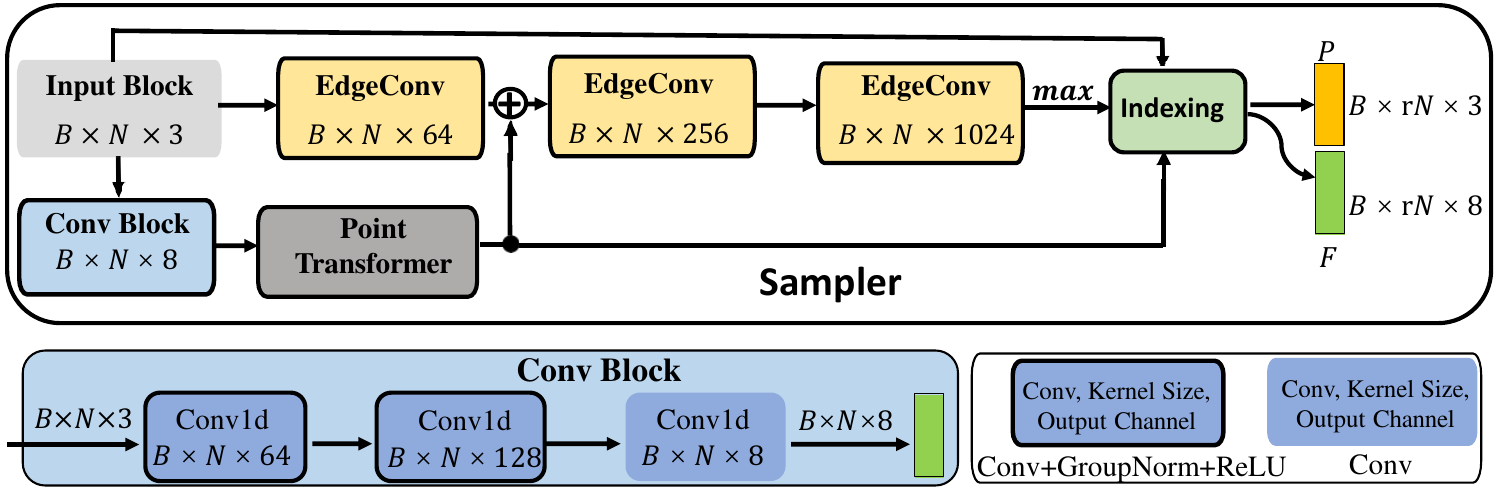}}
    \caption{The Encoder. $\oplus$ and $\circleddash$ denote concatenation and subtraction of tensor, respectively. Indexing means sampling points and features according to the index.} 
    \label{encoder}
    \end{center}
    \vskip -0.4in
\end{figure*}

\textbf{Architecture.} The overall framework of our \textbf{COT-PCC} for PCC is shown in Fig.~\ref{Framework}. As illustrated in the upper part, the generator consists of the encoder and decoder. The same \textbf{decoder} structure as DPCC~\cite{He_2022_CVPR} is applied for the implementation. For the \textbf{encoder}, notably, we introduce a learnable locally density-sensitive sampling module rather than the farthest point sampling~(FPS) in DPCC to let the network learn a local distribution-friendly sampling strategy, inspired by SampleNet~\cite{lang2020samplenet}. Specifically, the encoder down-sampled the point cloud in three stages, and the Sampler's parameters are shared in each stage. As shown in Fig.~\ref{encoder}, given the input $n$ points, \textbf{Sampler} first utilizes the Conv Block and Point Transformer~\cite{zhao2021point} to extract the features of the input points. Then, the Sampler utilizes 3 EdgeConv ~\cite{wang2019dynamic} layers to select $m=nr$ points $P_i$ ($i=1,2,3$) and the corresponding features $F_i$ with the most significant features according to the downsampling ratio $r$. Moreover, the Sampler accepts downsampling at any scale, thus enabling point cloud compression at any bitrate. The quantizers perform quantization compression for geometric coordinates and features based on entropy.

For the discriminator $J$ of Eq.\eqref{eq:PCCGAN}, it first uses residual structure and $1\times1$ convolution to extract features and then applies a 2-layer MLPs to identify original and reconstructed point clouds. More implementation details can be found in our open-source codes.

\vspace{-2mm}
\section{Experiments}
In this section, we evaluate our method by comparing it to state-of-the-art methods on compression rate and reconstruction accuracy.

\subsection{Setup}
\textbf{Datasets.}
We conduct experiments on SemanticKITTI ~\cite{behley2019semantickitti} and ShapeNet~\cite{chang2015shapenet}, consisting of LiDAR point clouds of outdoor scenes and sampled point clouds of CAD models. For a fair comparison, we utilize the same data preparation protocol of DPCC\cite{He_2022_CVPR}. All point clouds are first normalized to $100m^3$ cubes and divided into non-overlapping blocks of $12m^3$ and $22m^3$ for SemanticKITTI and ShapeNet respectively, while each block is further normalized to [-1, 1]. Then, we obtained the point clouds by non-uniform sampling. To further validate the generalization of the proposed method, we also directly apply the model trained on ShapeNet to the MPEG PCC dataset~\cite{MPEG}. 

\noindent\textbf{Baselines and metrics.}  We compare to state-of-the-art methods of both rule-based: Google Draco~\cite{galligan2018google}, MPEG Anchor~\cite{mekuria2016design}, G-PCC~\cite{graziosi2020overview}; and learning-based: Depeco~\cite{wiesmann2021deep}, PCGC~\cite{wang2021lossy}, DPCC, SparsePCGC~\cite{wang2022sparse}. For the qualitative evaluation, we demonstrate the comparison results with Google Draco, MPEG Anchor, and DPCC. For the quantitative analysis, we adopt the symmetric point-to-point Chamfer Distance (CD) and point-to-plane PSNR for geometry accuracy and Bits per Point~(Bpp) for compression rate.

\noindent\textbf{Implementation details.} We train the model for a total of 50 epochs and use the Adam optimizer with a learning rate of $0.0001$, and empirically set $\beta=100$, $\gamma=0.001$. Besides, to obtain different compressed bitrates, we vary the coefficient $\lambda$ of rate loss and choose an appropriate downsampling ratio $r$ from $\{1/2,1/3\}$ for the three stages of the encoder.  In the Sampler module, we first encode the point cloud into the feature space of 1024, then select the point with the largest value on each dimension to obtain the feature significance point set $S_1$, and then randomly sample $nr$ points from $S_1$ according to the ratio $r$. If $|S_1|$, which denotes the cardinality of $S_1$, is less than $nr$, we select the largest $k$ points on each dimension to construct the feature significance set $S_k$. In this case, $k$ is adaptively calculated by rounding up $nr/|S_1|$.

\vskip -0.15in
\subsection{Quantitative results}
\begin{figure}[htbp]
    \vspace{-2mm}
    \centering
    \includegraphics[width=1.0\linewidth]{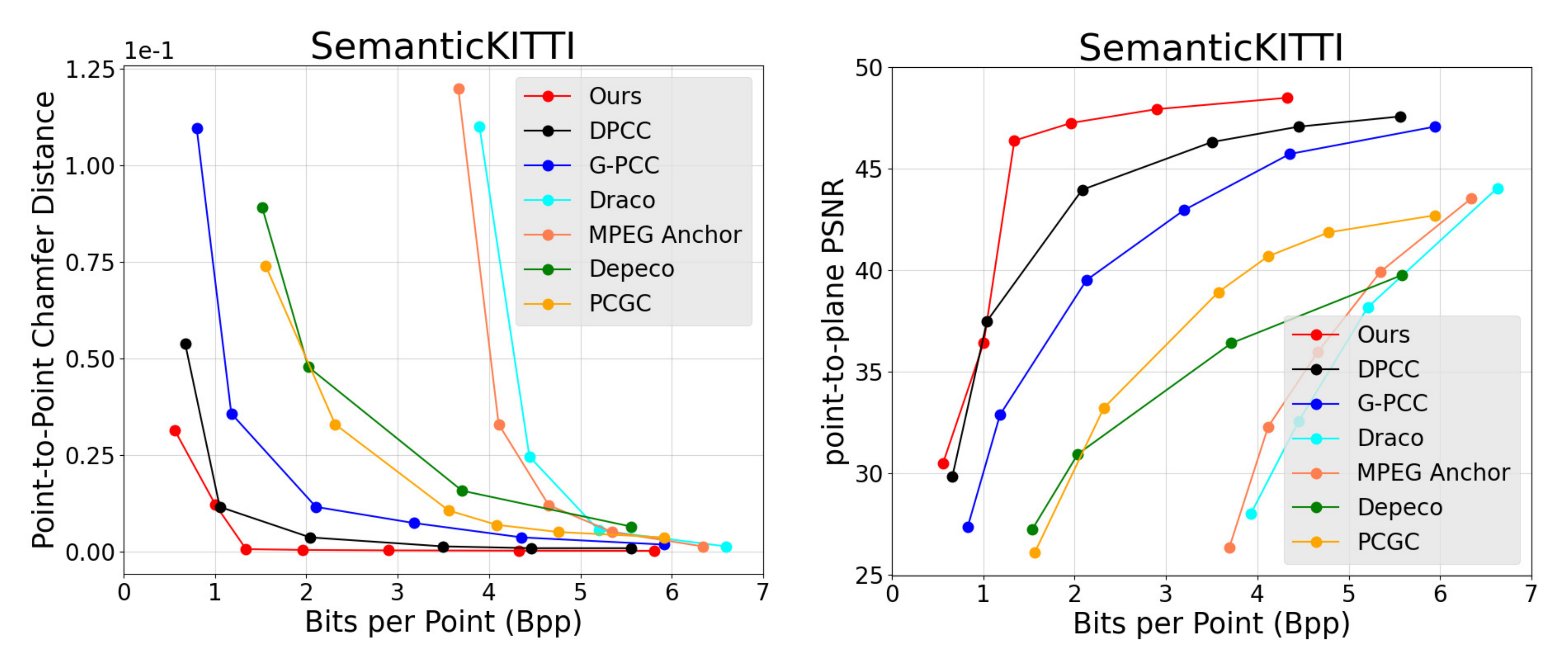}
    \includegraphics[width=1.0\linewidth]{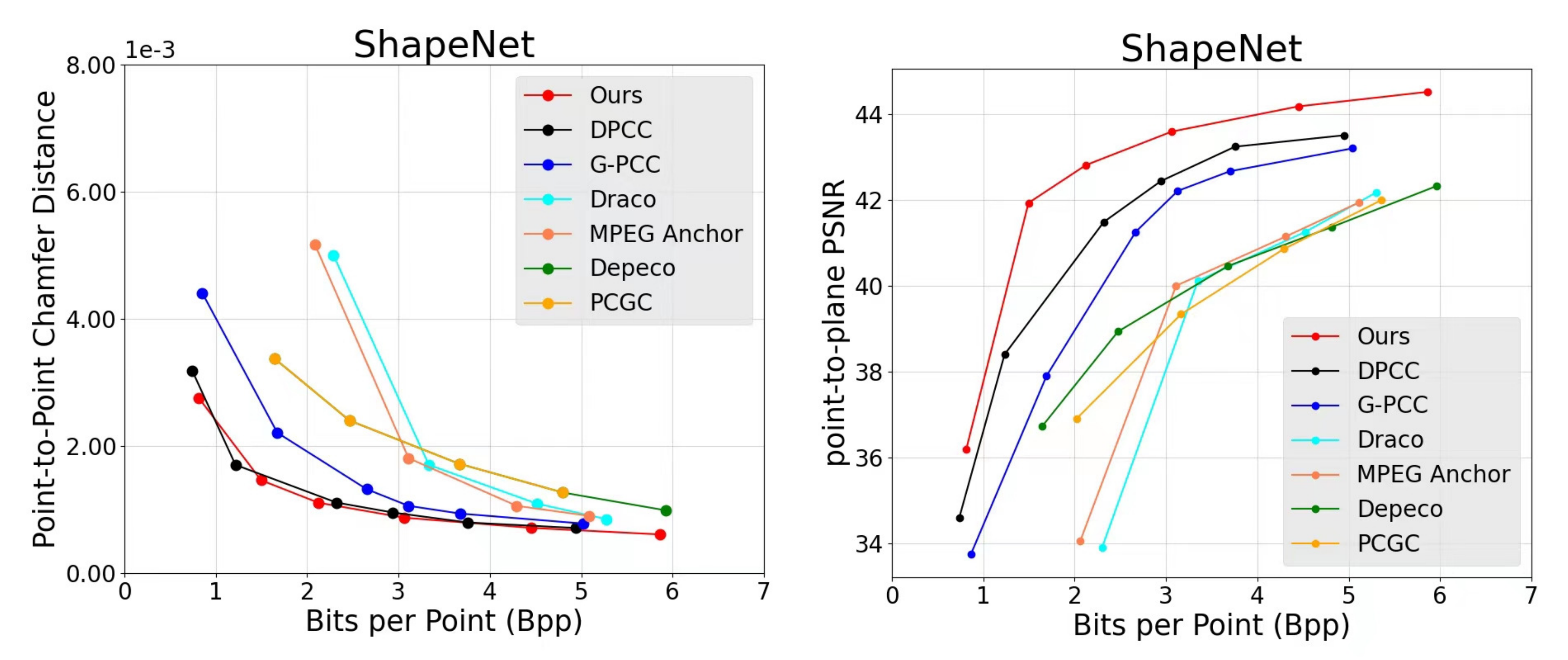}
        \vspace{-3mm}		
    \caption{Quantitative results on SemanticKITTI and ShapeNet.}
    \label{curve}
    \vspace{-5mm}		
\end{figure}
We first compare our method against state-of-the-art methods on the rate-distortion trade-off. In Fig.~\ref{curve}, we show the CD and PSNR of all methods against bits per point~(Bpp), where the results of all baselines are from DPCC~\cite{He_2022_CVPR}. The decompressed results of our model demonstrate lower CD loss and higher PSNR scores under the same bitrate constraint for both sparse point clouds of SemanticKITTI and dense ones of ShapeNet. It indicates that our COT-PCC yields better reconstruction performance consistently across the full spectrum of Bpp. In particular, the advantage of our method is more obvious on the PSNR metric, which reflects more global consistency. In addition, 
COT-PCC outperforms G-PCC by 5$\sim$6~dB in point-to-surface PSNR when $Bpp<3$, which shows a significant advantage in low Bpp scenarios.

\begin{figure}[htbp]
    \centering
    \includegraphics[width=0.49\linewidth]{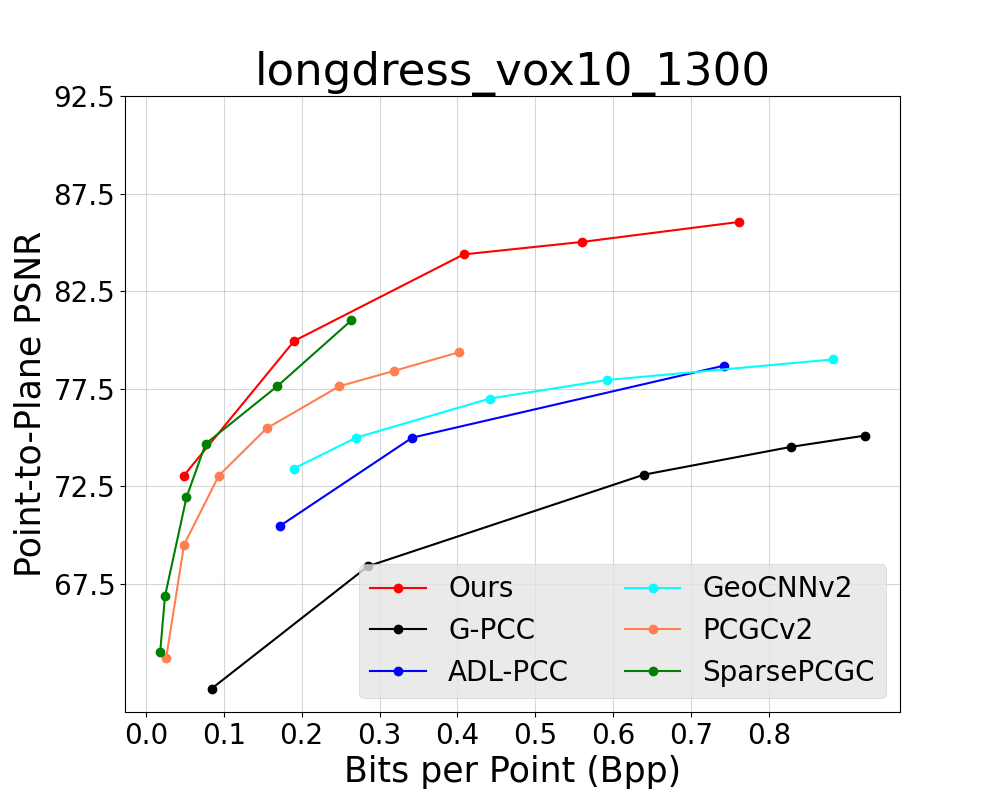}
    \includegraphics[width=0.49\linewidth]{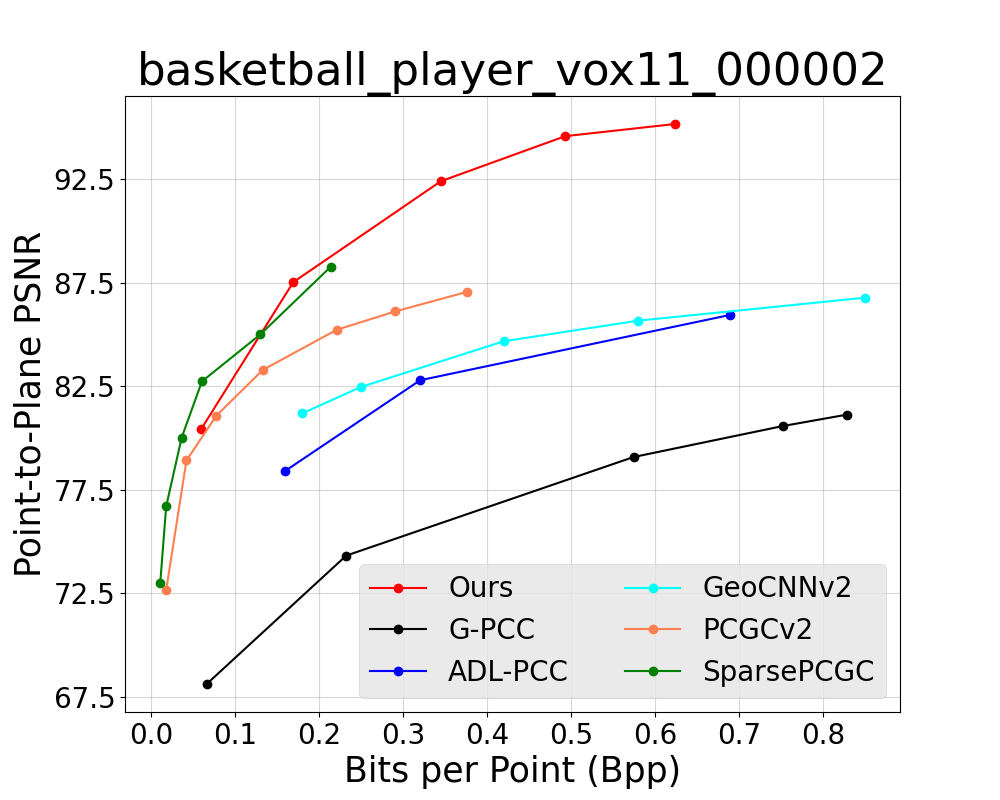}
        \vspace{-2mm}		
    \caption{Quantitative results on MPEG PCC dataset.}
    \label{fig:MPEG}
\end{figure}

To validate the generalization of the model, we directly apply the COT-PCC model trained on Shapenet to the MEPG dataset. The results are shown in Fig.~\ref{fig:MPEG}, where the results of other baselines are obtained from SparsePCGC~\cite{wang2022sparse}. It can be found that our method performs the best on ``longdress'' and ``basketball player'', indicating that our method has strong generalization ability.
\vspace{-2mm}
\vspace{-2mm}		

\subsection{Qualitative results}
\begin{figure}[htbp]
    \centering
    \includegraphics[width=1.0\linewidth]{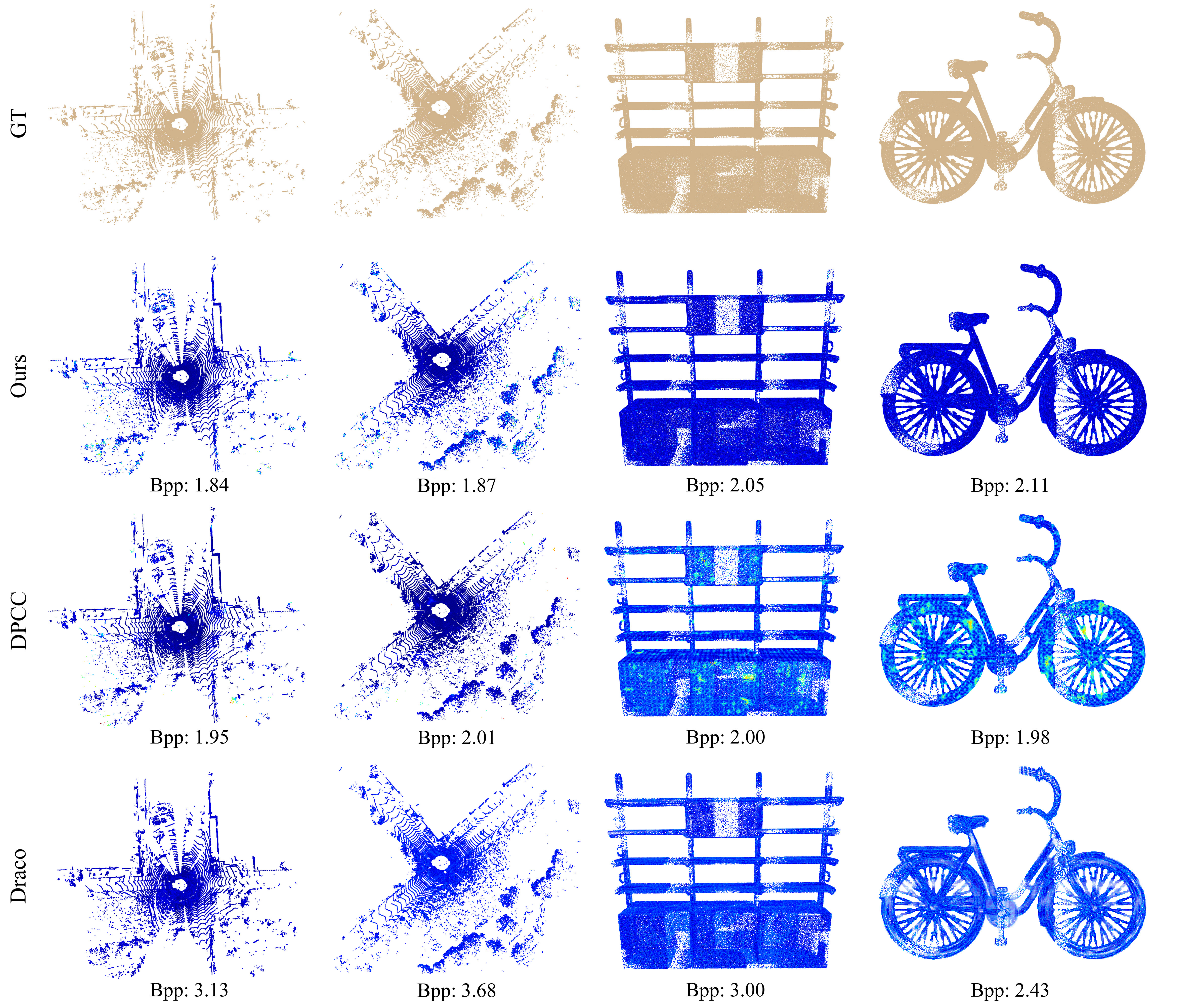}
    \includegraphics[width=1.0\linewidth]{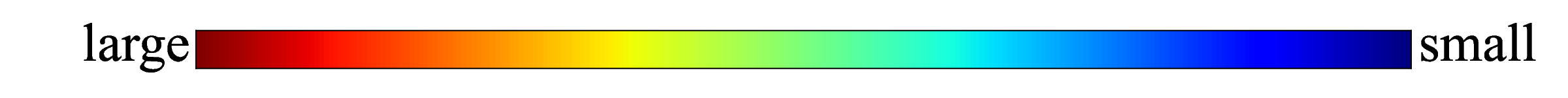}
        \vspace{-3mm}		
    \caption{Decompression examples from SemanticKITTI and ShapeNet datasets. Cool colors indicate small errors.}
    \label{erroMaps}
        \vspace{-8mm}		
\end{figure}

The errors of the reconstructed results by different methods with approximate bitrates are visualized in Fig.~\ref{erroMaps}. Among them, the first two data are from sparse and non-uniform SemanticKITTI, and the last two are from the dense CAD point clouds dataset, ShapeNet. 
As shown in Fig.~\ref{erroMaps}, Draco~\cite{galligan2018google} and MPEG~\cite{mekuria2016design} typically need a high Bpp to achieve a satisfactory reconstruction. Besides, due to the operation of voxelization, point clouds reconstructed by these two methods show obvious grid artifacts. 
DPCC~\cite{He_2022_CVPR} overly emphasizes local density uniformity which leads to large distortion in some high-frequency detail areas, such as the bicycle wheels. By optimizing the Wasserstein distance between the decompressed and the real point cloud distributions, our COT-PCC achieves the best performance in both global distribution and local detail, even with the lowest bitrates.

\subsection{Ablation Study}
\begin{figure}[htb!]
\vspace{-2mm}
    \centering
    \includegraphics[width=1.0\linewidth]{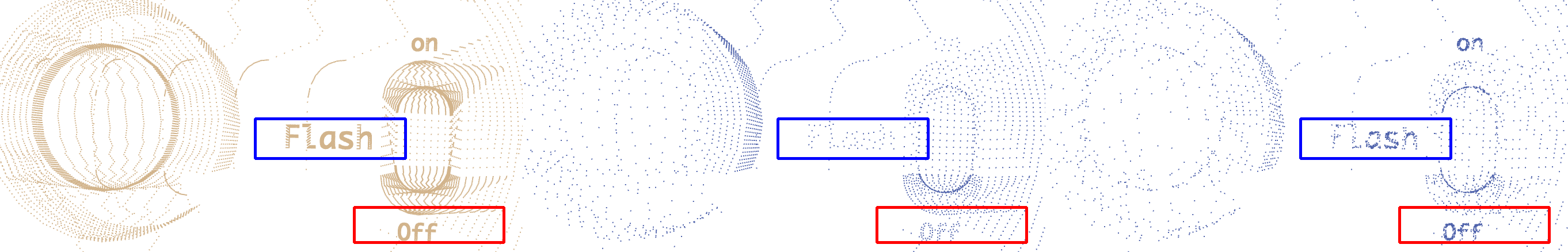}
    \includegraphics[width = 1.0\linewidth]{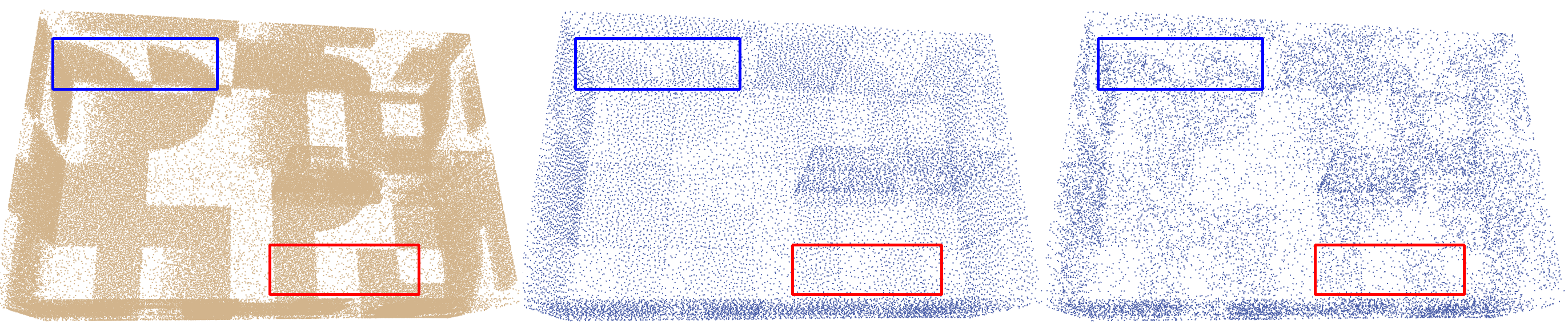}
    \subcaptionbox*{Input\&Target \qquad\qquad FPS \qquad\qquad\textbf{Sampler}}
    {\includegraphics[width = 1.0\linewidth]{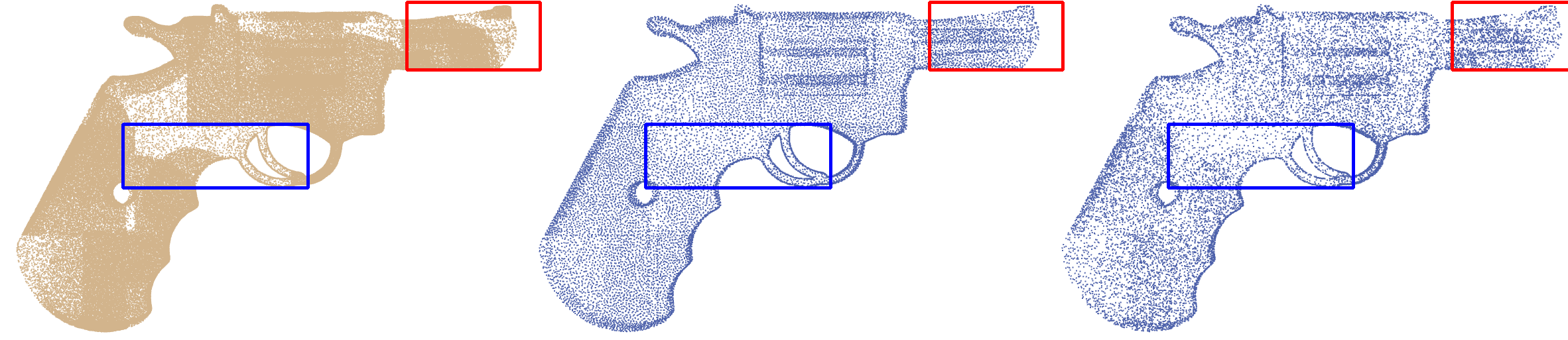}}
    \caption{Visualization of compressed point clouds.}
    \label{compression}
    \vspace{-2mm}
\end{figure}

To evaluate the effectiveness of our Sampler, we conduct an ablation study on SemanticKITTI and show it in Table ~\ref{tab:ablation}. Compared with FPS, our learnable \textbf{Sampler} achieves better results under the same Bpp constraint. The effectiveness of COT can be reflected in the comparison with DPCC, so no ablation experiment is designed here. Furthermore, we visualized the compressed point cloud results of PFS and our proposed Sampler. As shown in Fig.~\ref{compression}, we can see that FPS tends to obtain uniformly distributed compressed point clouds, and our learnable Sampler can better maintain original point cloud distribution and high-frequency details.
\begin{table}[ht!]
    \centering
    \caption{Ablation on Sampler.}
            \vspace{-2mm}		
    \begin{tabular}{l|c|c|c}
    \hline
    Model  & Bpp & CD$(10^{-3})$  & PSNR \\ \hline
    FPS	& 2.277 & 0.797 & 45.782 	 \\
    \textbf{Sampler}  & \textbf{1.963} &\textbf{0.676}& \textbf{46.383}\\
    \hline
    \end{tabular}
    \label{tab:ablation}  
    \vspace{-2mm}
\end{table}


\section{Conclusion}
In this paper, we proposed a COT-based framework for PCC. By re-formulating PCC as a bitrate-constrained OT problem, our method exploits the adversarial training for the PCC task by making the output distribution consistent with the real distribution from the global perspective, thus improving the global geometry quality and local high-frequency details consistency. We incorporated a quadratic Wasserstein for the stable training of GANs and introduced a learnable locally density-sensitive sampler to facilitate the downsampling stage of the compression process. We performed quantitative and qualitative experiments on sparse LiDAR and dense CAD point cloud data and validated the effectiveness of our method, especially in global geometry. 

\noindent\textbf{Limitations} The key limitation of the proposed method is that it requires an additional discriminator to assist in training the generator. In future research, we will investigate how to improve the efficiency of solving OT mapping.
	
	
\bibliographystyle{IEEEbib}
\small{\bibliography{icme}}

\begin{thebibliography}{10}

\bibitem{graziosi2020overview}
D~Graziosi, O~Nakagami, S~Kuma, A~Zaghetto, T~Suzuki, and A~Tabatabai,
\newblock ``An overview of ongoing point cloud compression standardization
  activities: Video-based (v-pcc) and geometry-based (g-pcc),''
\newblock {\em APSIPA Trans. Signal Inf. Process.}, vol. 9, 2020.

\bibitem{zhang2022novel}
Wei Zhang, Youguang Yu, and Fuzheng Yang,
\newblock ``A novel grid-based geometry compression framework for spinning
  lidar point clouds,''
\newblock in {\em ICME}. IEEE, 2022, pp. 1--6.

\bibitem{huang2020octsqueeze}
Lila Huang, Shenlong Wang, Kelvin Wong, Jerry Liu, and Raquel Urtasun,
\newblock ``Octsqueeze: Octree-structured entropy model for lidar
  compression,''
\newblock in {\em CVPR}, 2020, pp. 1313--1323.

\bibitem{quach2019learning}
Maurice Quach, Giuseppe Valenzise, and Frederic Dufaux,
\newblock ``Learning convolutional transforms for lossy point cloud geometry
  compression,''
\newblock in {\em ICIP}. IEEE, 2019, pp. 4320--4324.

\bibitem{wang2021lossy}
Jianqiang Wang, Hao Zhu, Haojie Liu, and Zhan Ma,
\newblock ``Lossy point cloud geometry compression via end-to-end learning,''
\newblock {\em TCSVT}, vol. 31, pp. 4909--4923, 2021.

\bibitem{quach2020improved}
Maurice Quach, Giuseppe Valenzise, and Frederic Dufaux,
\newblock ``Improved deep point cloud geometry compression,''
\newblock in {\em MMSP}. IEEE, 2020, pp. 1--6.

\bibitem{wang2022sparse}
Jianqiang Wang, Dandan Ding, Zhu Li, Feng, et~al.,
\newblock ``Sparse tensor-based multiscale representation for point cloud
  geometry compression,''
\newblock {\em TPAMI}, 2022.

\bibitem{song2023efficient}
Rui Song, Chunyang Fu, Shan Liu, and Ge~Li,
\newblock ``Efficient hierarchical entropy model for learned point cloud
  compression,''
\newblock in {\em CVPR}, 2023, pp. 14368--14377.

\bibitem{liu2023pchm}
Lei Liu, Zhihao Hu, and Jing Zhang,
\newblock ``Pchm-net: A new point cloud compression framework for both human
  vision and machine vision,''
\newblock in {\em ICME}. IEEE, 2023, pp. 1997--2002.

\bibitem{huang20193d}
Tianxin Huang and Yong Liu,
\newblock ``3d point cloud geometry compression on deep learning,''
\newblock in {\em Proceedings of the 27th ACM MM}, 2019, pp. 890--898.

\bibitem{yan2019deep}
Wei Yan, Shan Liu, Thomas~H Li, Zhu Li, Ge~Li, et~al.,
\newblock ``Deep autoencoder-based lossy geometry compression for point
  clouds,''
\newblock {\em arXiv preprint:1905.03691}, 2019.

\bibitem{gu20203d}
Shuai Gu, Junhui Hou, Huanqiang Zeng, and Hui Yuan,
\newblock ``3d point cloud attribute compression via graph prediction,''
\newblock {\em IEEE SPL}, vol. 27, pp. 176--180, 2020.

\bibitem{wiesmann2021deep}
Louis Wiesmann, Andres Milioto, Xieyuanli Chen, Cyrill Stachniss, and Jens
  Behley,
\newblock ``Deep compression for dense point cloud maps,''
\newblock {\em Robot. Autom. Lett.}, vol. 6, pp. 2060--2067, 2021.

\bibitem{wang2021multiscale}
Jianqiang Wang, Dandan Ding, Zhu Li, and Zhan Ma,
\newblock ``Multiscale point cloud geometry compression,''
\newblock in {\em DCC}. IEEE, 2021, pp. 73--82.

\bibitem{He_2022_CVPR}
Yun He, Xinlin Ren, Danhang Tang, Yinda Zhang, Xiangyang Xue, and Yanwei Fu,
\newblock ``Density-preserving deep point cloud compression,''
\newblock in {\em CVPR}, 2022, pp. 333--342.

\bibitem{gao2023point}
Pan Gao, Lijuan Zhang, Lei Lei, and Wei Xiang,
\newblock ``Point cloud compression based on joint optimization of graph
  transform and entropy coding for efficient data broadcasting,''
\newblock {\em IEEE Trans. Broadcast.}, 2023.

\bibitem{que2021voxelcontext}
Zizheng Que, Guo Lu, and Dong Xu,
\newblock ``Voxelcontext-net: An octree based framework for point cloud
  compression,''
\newblock in {\em CVPR}, 2021, pp. 6042--6051.

\bibitem{2019Rethinking}
Y.~{Blau} and T.~{Michaeli},
\newblock ``Rethinking lossy compression: The rate-distortion-perception
  tradeoff,''
\newblock in {\em ICML}, 2019, pp. 675--685.

\bibitem{yan2021perceptual}
Zeyu Yan, Fei Wen, Rendong Ying, Chao Ma, and Peilin Liu,
\newblock ``On perceptual lossy compression: The cost of perceptual
  reconstruction and an optimal training framework,''
\newblock in {\em ICML}, 2021, pp. 11682--11692.

\bibitem{zhang2021universal}
George Zhang, Jingjing Qian, Jun Chen, and Ashish Khisti,
\newblock ``Universal rate-distortion-perception representations for lossy
  compression,''
\newblock {\em NeurIPS}, vol. 34, pp. 11517--11529, 2021.

\bibitem{de2016compression}
Ricardo~L De~Queiroz and Philip~A Chou,
\newblock ``Compression of 3d point clouds using a region-adaptive hierarchical
  transform,''
\newblock {\em TIP}, vol. 25, no. 8, pp. 3947--3956, 2016.

\bibitem{biswas2020muscle}
Sourav Biswas, Jerry Liu, Kelvin Wong, Shenlong Wang, and Raquel Urtasun,
\newblock ``Muscle: Multi sweep compression of lidar using deep entropy
  models,''
\newblock {\em NeurIPS}, vol. 33, pp. 22170--22181, 2020.

\bibitem{guarda2020adaptive}
Andr{\'e}~FR Guarda, Nuno~MM Rodrigues, and Fernando Pereira,
\newblock ``Adaptive deep learning-based point cloud geometry coding,''
\newblock {\em J Sel Top Signal Process}, vol. 15, pp. 415--430, 2020.

\bibitem{thomas2006elements}
MTCAJ Thomas and A~Thomas Joy,
\newblock {\em Elements of information theory},
\newblock Wiley-Interscience, 2006.

\bibitem{shannon1959coding}
Claude~E Shannon et~al.,
\newblock ``Coding theorems for a discrete source with a fidelity criterion,''
\newblock {\em IRE Nat. Conv. Rec}, vol. 4, no. 142-163, pp. 1, 1959.

\bibitem{balle2016end}
Johannes Ball{\'e}, Valero Laparra, and Eero~P Simoncelli,
\newblock ``End-to-end optimized image compression,''
\newblock {\em arXiv preprint arXiv:1611.01704}, 2016.

\bibitem{liu2021lossy}
Huan Liu, George Zhang, Jun Chen, and Ashish~J Khisti,
\newblock ``Lossy compression with distribution shift as entropy constrained
  optimal transport,''
\newblock in {\em ICLR}, 2021.

\bibitem{wen2020lossy}
Xuanzheng Wen, Xu~Wang, Junhui Hou, Lin Ma, Yu~Zhou, and Jianmin Jiang,
\newblock ``Lossy geometry compression of 3d point cloud data via an adaptive
  octree-guided network,''
\newblock in {\em ICME}. IEEE, 2020, pp. 1--6.

\bibitem{peyre2019computational}
Gabriel Peyr{\'e}, Marco Cuturi, et~al.,
\newblock ``Computational optimal transport: With applications to data
  science,''
\newblock {\em Found. Trends Mach. Learn.}, pp. 355--607, 2019.

\bibitem{li2022weakly}
Zezeng Li, Weimin Wang, Na~Lei, and Rui Wang,
\newblock ``Weakly supervised point cloud upsampling via optimal transport,''
\newblock in {\em ICASSP 2022-2022}. IEEE, 2022, pp. 2564--2568.

\bibitem{monge1781memoire}
Gaspard Monge,
\newblock ``M{\'e}moire sur la th{\'e}orie des d{\'e}blais et des remblais,''
\newblock {\em Histoire de l'Acad{\'e}mie Royale des Sciences de Paris}, 1781.

\bibitem{lei2019geometric}
Na~Lei, Kehua Su, Li~Cui, Shing-Tung Yau, and Xianfeng~David Gu,
\newblock ``A geometric view of optimal transportation and generative model,''
\newblock {\em CAGD}, vol. 68, pp. 1--21, 2019.

\bibitem{li2019pu}
Ruihui Li, Xianzhi Li, Chi-Wing Fu, Daniel Cohen-Or, and Pheng-Ann Heng,
\newblock ``Pu-gan: a point cloud upsampling adversarial network,''
\newblock in {\em ICCV}, 2019, pp. 73--82.

\bibitem{2018Unreasonable}
R.~{Zhang}, P.~{Isola}, A.~A. {Efros}, E.~{Shechtman}, and O.~{Wang},
\newblock ``The unreasonable effectiveness of deep features as a perceptual
  metric,''
\newblock in {\em CVPR}, 2018, pp. 586--595.

\bibitem{wganqc}
Huidong Liu, Xianfeng Gu, and Dimitris Samaras,
\newblock ``Wasserstein gan with quadratic transport cost,''
\newblock in {\em ICCV}, 2019, pp. 4832--4841.

\bibitem{lang2020samplenet}
Itai Lang, Asaf Manor, and Shai Avidan,
\newblock ``Samplenet: Differentiable point cloud sampling,''
\newblock in {\em CVPR}, 2020, pp. 7578--7588.

\bibitem{zhao2021point}
Hengshuang Zhao, Li~Jiang, Jiaya Jia, Philip~HS Torr, and Vladlen Koltun,
\newblock ``Point transformer,''
\newblock in {\em ICCV}, 2021, pp. 16259--16268.

\bibitem{wang2019dynamic}
Yue Wang, Yongbin Sun, Ziwei Liu, Sanjay~E Sarma, Michael~M Bronstein, and
  Justin~M Solomon,
\newblock ``Dynamic graph cnn for learning on point clouds,''
\newblock {\em ACM TOG}, vol. 38, no. 5, pp. 1--12, 2019.

\bibitem{behley2019semantickitti}
Jens Behley, Martin Garbade, Andres Milioto, Jan Quenzel, Sven Behnke, Cyrill
  Stachniss, and Jurgen Gall,
\newblock ``Semantickitti: A dataset for semantic scene understanding of lidar
  sequences,''
\newblock in {\em ICCV}, 2019, pp. 9297--9307.

\bibitem{chang2015shapenet}
Angel~X Chang, Thomas Funkhouser, Guibas, et~al.,
\newblock ``Shapenet: An information-rich 3d model repository,''
\newblock {\em arXiv preprint arXiv:1512.03012}, 2015.

\bibitem{MPEG}
``Mpeg pcc dataset,'' Accessed: 2022.

\bibitem{galligan2018google}
Frank Galligan, Michael Hemmer, Ondrej Stava, Zhang, and Jamieson others,
\newblock ``Google/draco: a library for compressing and decompressing 3d
  geometric meshes and point clouds,'' 2018.

\bibitem{mekuria2016design}
Rufael Mekuria, Kees Blom, and Pablo Cesar,
\newblock ``Design, implementation, and evaluation of a point cloud codec for
  tele-immersive video,''
\newblock {\em TCSVT}, vol. 27, no. 4, pp. 828--842, 2016.

\end{thebibliography}

\end{document}